\documentclass{article}

\usepackage{PRIMEarxiv}

\usepackage[utf8]{inputenc} 
\usepackage[T1]{fontenc}    
\usepackage{hyperref}       
\usepackage{url}            
\usepackage{booktabs}       
\usepackage{amsfonts}       
\usepackage{nicefrac}       
\usepackage{microtype}      
\usepackage{lipsum}
\usepackage{fancyhdr}       
\usepackage{graphicx}       
\usepackage{amsmath}
\usepackage{tabularx}
\usepackage{authblk}
\usepackage{array} 
\usepackage{makecell} 
\graphicspath{{media/}}     

\begin{document}  
\title{Person Recognition using Facial Micro-Expressions with Deep Learning
}

\author{
  Tuval Kay*, Yuval Ringel*, Khen Cohen , Mor-Avi Azulay , David Mendlovic\textsuperscript{\textdagger} \\
  Electrical Engineering, Computer Vision and Optics Lab \\
  Tel Aviv University \\
  \texttt{tuvalkay@mail.tau.ac.il, yuvalringel@mail.tau.ac.il} \\

}

\maketitle
{\footnotesize \textsuperscript{*}These authors contributed equally to this work.}

\begin{abstract}
This study investigates the efficacy of facial micro-expressions as a soft biometric for enhancing person recognition, aiming to broaden the understanding of the subject and its potential applications. We propose a deep learning approach designed to capture spatial semantics and motion at a fine temporal resolution. Experiments on three widely-used micro-expression databases demonstrate a notable increase in identification accuracy compared to existing benchmarks, highlighting the potential of integrating facial micro-expressions for improved person recognition across various fields.
\end{abstract}

\section{Introduction}
Face recognition, the process of identifying individuals based on their facial characteristics, is an innate ability for humans and has become one of the most reliable and well-studied biometrics in automatic person recognition research. However, conventional face recognition approaches can suffer significantly in unconstrained environments. To address these challenges, researchers have explored the integration of appearance-based soft biometric characteristics extracted from facial images, such as facial marks, skin color, and hairstyle \cite{becerra2019}. Additionally, behavior-based facial soft biometrics, including head dynamics, visual speech \cite{saeed2006}, and facial expressions, have been investigated (section 2).

Facial expressions were initially considered a source of variation in face recognition, but recent research (section 2) has demonstrated their potential for improving person recognition results when incorporated as a soft biometric. These expressions can be broadly classified into macro and micro-expressions. Macro-expressions are the six commonly recognized facial expressions—anger, fear, disgust, surprise, sadness, and joy—typically lasting between 0.5 to 4 seconds \cite{ekman2007}. In contrast, micro-expressions are rapid (0.04 to 0.2 seconds) and subtle, localized to specific areas of the face \cite{ekman2009}. Due to their fleeting and nuanced nature, even trained professionals can only achieve a 47\% recognition rate for micro-expressions \cite{frank2009}.

While most existing studies on facial expressions as soft biometrics focus on macro-expressions, their use for person recognition is limited by the potential for posing or faking \cite{saeed2021}. In contrast, micro-expressions, with their spontaneous, rapid, and subtle characteristics, offer a promising avenue for advancing person recognition and combating spoofing attacks. Consequently, this paper specifically investigates the potential of micro-expressions as a soft biometric for person recognition.

Given the widespread success of deep learning models in various computer vision tasks, we aim to explore their capacity to leverage facial micro-expressions for enhanced person recognition. Our hypothesis is that a convolutional neural network (CNN) can identify and learn unique movement patterns of facial micro-expressions for each individual, thereby improving recognition performance. However, a major challenge associated with deep learning models is their "black box" nature, which obscures the understanding of factors influencing the network's learning. To address this issue and validate our hypothesis, we visualize and analyze what the model has learned from facial micro-expressions during the recognition process (section 4).

\section{Literature review}
In this section, we review previous studies that explore facial expressions and soft biometrics for person recognition. We first discuss research that employs traditional feature extraction techniques, followed by studies using deep learning models. Finally, we highlight the gap in the literature and the motivation for our study.

\subsection{Traditional feature extraction techniques}
In a study by \cite{saeed2021}, the authors proposed a method that combines traditional facial features represented by Local Gabor Binary Pattern (LGBP) with soft biometrics, specifically micro-expressions. Two feature extraction methods commonly used for video-based micro-expression recognition were employed: Local Binary Patterns from Three Orthogonal Planes (LBP-TOP) and Fuzzy Histogram of Optical Flow Orientations (FHOFO). The classification was achieved using Support Vector Machines (SVM).

In \cite{lee2017}, the authors explored the use of three soft biometric traits: race, gender, and emotion. After normalizing the videos containing micro-expressions, they applied Multimodal Discriminant Analysis to decompose them into race, gender, and emotion components. The LBP-TOP feature vector and SVM classifier were then used for person and emotion recognition. Experiments on the SMIC database showed that increasing all components simultaneously reduced person recognition results, while increasing only race and gender components decreased emotion recognition results.
\cite{gaweda2011} proposed a system based on facial dynamics during facial expressions for person recognition. The feature extraction was based on Active Appearance Model (AAM) parameters, and classification was achieved using Hidden Markov Models (HMM).
In \cite{gavrilescu2016}, the authors proposed using facial macro-expression as a soft biometric. The system consists of two blocks: one for person recognition using Principal Component Analysis (PCA) and the other for expression recognition using an Artificial Neural Network (ANN). By merging the results from both blocks using a decision tree.

\subsection{Deep learning models}
\cite{hammer2018} investigated using changes in facial expressions for person recognition, with experiments conducted on a self-collected database of 61 individuals who performed six facial expressions. Two feature extraction techniques were employed: a fine-tuned VGG-CNN and geometric feature extraction from facial landmarks, with LSTM used for classification.
An expression invariant face recognition system was proposed in \cite{peng2018}. By applying expression manifolds to generate synthetic expressions from neutral faces, the classification accuracy improved significantly. 
Finally, in \cite{shreve2016}, action units were extracted using FaceReader and compared using distance measures. Experiments on a self-collected database of 96 individuals participating in a game-show type quiz with a prize incentive.

\subsection{Gap in the literature and motivation}
Although the studies mentioned above have demonstrated promising results, they also exhibit certain limitations. Many of them have relied on self-collected databases, making it difficult to compare results. More importantly, most studies have focused on facial macro-expressions and have not explored the potential of micro-expressions to further improve person recognition. Future research should consider addressing these limitations and delve deeper into the potential of micro-expressions as a soft biometric for person recognition using deep learning models.

\begin{figure}
    \centering 
    \includegraphics[width=0.6\textwidth]{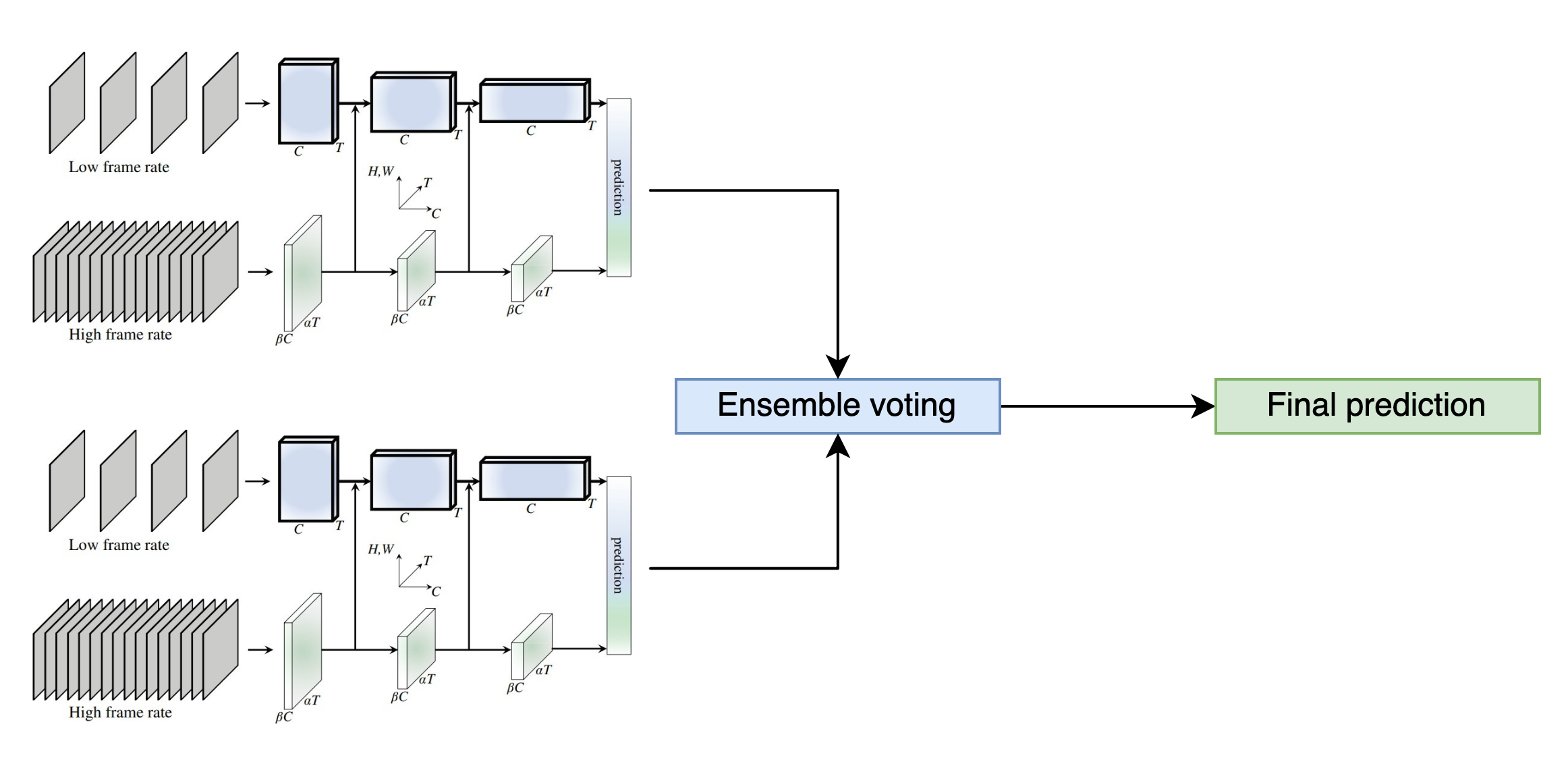}    
    \caption{Ensemble of SlowFast CNN models with distinct hyperparameters for micro-expression-based person recognition, illustrating input video sequence processing, individual predictions, and final person identity determined through ensemble voting.} 
    \label{fig_mom0}%
\end{figure}

\section{Proposed Method}
\
In this paper, we address the problem of micro-expression-based person recognition as a soft biometric, which poses unique challenges due to the subtle and involuntary nature of these facial expressions. To tackle this problem, we propose the use of the SlowFast CNN model, an architecture specifically designed for capturing spatial semantics and motion at fine temporal resolution. The SlowFast CNN consists of two separate pathways: the slow pathway focuses on high-level semantic content by processing a small number of video frames, while the fast pathway captures fine-grained motion information by processing a larger number of frames. These pathways are combined to produce a single output that benefits from both the high-level semantic understanding and the detailed motion information \cite{feichtenhofer2019}.

Our choice of the SlowFast CNN model is motivated by its impressive performance in capturing rapid human actions, such as clapping hands. We found that the model's cross-stream fusion enables learning of true spatiotemporal features, rather than merely separate appearance and motion features. Additionally, the networks can learn both highly class-specific local representations and generic representations that serve a range of classes \cite{feichtenhofer2018}. We hypothesized that this unique combination of features allows for optimal utilization of micro-movements occurring during micro-expressions for facial recognition.

To better suit the problem of micro-expression analysis, we adapted the SlowFast CNN model by employing an Ensemble Voting approach with multiple SlowFast networks, each having different hyperparameters. This adaptation allows for more robust and accurate recognition by leveraging the complementary strengths of each individual network, thus improving the overall performance of the system.

To the best of our knowledge, this approach has not been previously explored.

\subsection{Preprocess}
First, We cropped the videos around the faces to dispose any noise and unwanted interference. Next, each frame was resized to match the dimensions of the data in \cite{saeed2021}. SMIC frames dimensions were 150X150, CASME2’s were 300X300 and SAMM’s were 400X400 (section 4). For each video in the dataset, we extracted a “window” of \(W=64\)  frames around the apex of the micro-expression. in order to use all the available data, if a clip had less frames than \(W\), we duplicated the first and last frames until the total was \(W\) (Since there is no movement except the micro-expression, the padding could not corrupt the results). The data was finally splitted into train and test sets with the same proportions of 50\% each, the same proportions as \cite{saeed2021}.

\section{Experiments}
In this section, we explain the employed databases, experimental setup, the conducted experiments, and the results obtained.
\subsection{Databases}
The experiments were performed on three facial micro-expression databases. These databases were selected because the image sequences include subject IDs, which are crucial for person recognition. Furthermore, these databases were recorded at high frame rates, which is necessary for studying short-duration micro-expressions.
The Spontaneous Micro-expression Database (SMIC)\cite{li2013}, The Spontaneous Actions and Micro-Movements (SAMM)\cite{davison2018}, and The Chinese Academy of Sciences Micro-expression (CASME) II \cite{yan2014}.
\subsection{Performance Metrics}
As our experiments are conducted in identification mode, we present the results using a commonly employed metric for accuracy i.e., Rank 1 identification accuracy. The identification rate is the percentage of the test samples that are correctly identified by the system at rank-1. In other words, it measures the proportion of test samples where the best match, based on the highest similarity score, is the correct one.

\begin{equation}
\text{Rank-1 probability} = \frac{n_{h}}{n_{t}} \times 100\%
\end{equation}

where $n_{h}$ refers to the number of test samples with the highest similarity score in the correct class, and $n_{t}$ represents the total number of test samples.

\subsection{Results}
\subsubsection{Single SlowFast performance}
In the first experiment, we evaluated the performance of a single SlowFast CNN using hyperparameters selected through grid search,in the SlowFast CNN model, the hyperparameters \(\alpha\) and \(\beta\) control the temporal downsampling factors for the slow and fast pathways, respectively. \(\alpha\) determines the number of input frames for the slow pathway, while \(\beta\) determines the ratio of input frames between the fast and slow pathways. Here, we present the hyperparameter values obtained through a grid search for each database as listed in Table 1. The results of this evaluation are presented in Table 2.
\begin{table}[htbp]
  \centering
  \caption{hyper parameters}
    \begin{tabular}{|c|c|c|c|}
    \hline
    Dataset & SAMM & CASME II & SMIC \\
    \hline
    \(\alpha\) & 4 & 4 & 16 \\
    \hline
    \(\beta\)  & $\frac{1}{8}$ & $\frac{1}{8}$ & $\frac{1}{16}$\\
    \hline
    Solver & adam & adam & adam\\
    \hline
    Learning Rate & 0.001 & 0.001 & 0.001\\
    \hline
    Batch Size & 16 & 16 & 32\\
    \hline
    \end{tabular}%
  \label{tab:table_label}%
\end{table}%

\begin{table}[htbp]
  \centering
  \caption{Rank 1 identification accuracy \%}
    \begin{tabular}{|c|c|c|c|}
    \hline
    Model & SAMM & CASME II & SMIC \\
    \hline
    SlowFast & 89.61\% & 87.4\% & 93.67\% \\
    \hline
    \end{tabular}%
  \label{tab:accuracy_table}%
\end{table}%
\subsubsection{Ensemble Voting Using multiple SlowFast networks}

We also investigated the effectiveness of an ensemble voting approach to further enhance our results. After generating predictions using various combinations of hyperparameters, we employed ensemble voting to pair different permutations of SlowFast networks, resulting in improved accuracy. For example, when comparing two SlowFast models with different solvers (Adam and AdamW) executed on the SMIC database using identical hyperparameter values (\(\alpha=16\), \(\beta=1\), Batch Size=32, Learning Rate=0.001), our ensemble approach yielded higher accuracy. Table 3 presents the improvements on SMIC results

\begin{table}[htbp]
  \centering
  \caption{Rank 1 identification accuracy \%}
    \begin{tabular}{|c|c|c|c|}
    \hline
    Model & adam & adamw & Ensemble of the two \\
    \hline
     SlowFast & 93.7\% & 89.9\% & 94.9\% \\
    \hline
    \end{tabular}%
  \label{tab:accuracy_table2}%
\end{table}%

The results demonstrate the effectiveness of the ensemble voting approach in improving the accuracy of our predictions. By combining the predictions from different models, we were able to achieve a higher level of accuracy than what could be achieved by individual models.

\subsection{Comparisons with Other techniques}
Among the techniques outlined in this study, only one prior work \cite{saeed2021} has investigated the use of micro-expressions for person recognition, while all other approaches have focused on macro-expressions. In our proposed approach, we apply the Ensemble Voting Using multiple SlowFast networks for the SMIC database and a single slowfast for SAMM, and CASME II databases to leverage the benefits of micro-expressions, resulting in an improved identification rate at rank-1 of 94.9\% , 89.61\%, and 87.4\% respectively. Notably, our proposed method outperforms the current benchmark set by \cite{saeed2021}.

\begin{table*}
  \centering
  \caption{Comparison of accuracy achieved by different methods}
  \label{tab:accuracy}
  \small
  \begin{tabular}{|c|c|c|c|c|}
    \hline
    Ref. & Database & Features & Classifier & Accuracy \\
    \hline
    [6] & SMIC, SAMM, CASME II & LGBP, LBP-TOP, FHOHO & SVM & 95.55\%, 74.85\%, 71.48\% \\
    \hline
    [8] & self-collected & AAM parameters & HMM & 90\% \\
    \hline
    [9] & \makecell{Honda/UCSD and \\youtube faces DB} & Facial images & PCA & 94.5\% \\
    \hline
    [10] & self-collected & VGG-face CNN and geometric features & LSTM & 96.2\% \\
    \hline
    [11] & \makecell{CK+, AR, MUG, JAFFE, \\MM1, Bosphorus} & AAM parameters & LDA & 99.9\% \\
    \hline
    [12] & self-collected & AU features & Distance measures & 85\% \\
    \hline
    [Ours] & SMIC, SAMM, CASME II & SlowFast CNN & \makecell{fully connected layer \\ with softmax } & 94.95\%, 89.61\%, 87.4\% \\
    \hline
  \end{tabular}
\end{table*}

\subsection{Challenges and Limitations}
One of the primary challenges in our study was obtaining a sufficient amount of labeled data for micro-expression recognition. High frame-rate videos are necessary to capture the subtle and rapid changes in facial features during micro-expressions. Annotating these videos is a time-consuming and labor-intensive task, as it requires experts who can accurately identify and tag the micro-expressions.

In the context of real-time applications, this challenge is further magnified. Developing a real-time system would require not only obtaining an extensive amount of labeled data for training but also implementing efficient algorithms that can process high frame-rate videos rapidly. The need for high-quality labeled data and the computational demands of real-time processing pose significant limitations to the current state of micro-expression recognition research.

Future work should focus on addressing these limitations by exploring alternative sources of labeled data, such as data augmentation techniques or leveraging unsupervised or semi-supervised learning methods. Additionally, optimizing the computational efficiency of recognition algorithms could facilitate real-time processing, thereby expanding the potential applications of micro-expression recognition for person identification.

\subsection{Neural Network Visualization Using Grad-CAM}
\cite{selvaraju2019} proposed a technique for producing ‘visual explanations’ for decisions from a large class of Convolutional Neural Network (CNN)-based models, making them more transparent and explainable. Gradient-weighted Class Activation Mapping (Grad-CAM) uses the gradients of any target concept (say ‘dog’ in a classification network or a sequence of words in captioning network) flowing into the final convolutional layer to produce a coarse localization map highlighting the important regions in the image for predicting the concept. As we can see in the Grad-CAM results (figure 2), the network has learned significant features from regions while micro-expressions occurred, which helps us demonstrate that the network has utilized the micro-expressions in the classification decision.

\begin{figure}[!htbp]
	\centering 
	\includegraphics[width=0.3\textwidth]{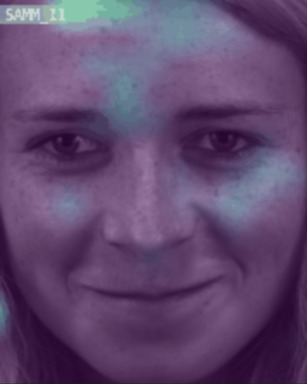}	
	\caption{Green highlights represent features learned by the neural network. Here, they correspond to the micro-movement being done by the subject.}
	\label{Neural Network Visualization Using Grad-Cam}
\end{figure}

\section{Summary and conclusions}
In this paper, we have investigated the potential of using facial micro-expressions as a soft biometric identifier. Although the publicly available micro-expression databases employed in this study are suitable for detection and recognition, their limited number of participants and small size pose challenges for person recognition and deep learning techniques. To address these challenges, we proposed an Ensemble Voting approach using multiple SlowFast CNN networks with different hyperparameters, demonstrating promising results.

Our experiments showed that the ensemble-based SlowFast CNN approach achieved recognition accuracies comparable to macro-expression-based techniques, highlighting the feasibility of micro-expressions for person recognition. In the future, we plan to explore the application of micro-expressions in adversarial settings, specifically spoofing attacks. The spontaneous, rapid, and subtle nature of micro-expressions makes them difficult to reproduce, potentially enabling detection and protection against deep fake abuse.

As larger datasets become available, further exploration of deep learning methods, such as our ensemble-based SlowFast CNN approach, can advance the field of micro-expression-based person recognition. We hope this work inspires continued investigation into the potential of micro-expressions as a soft biometric identifier and contributes to the development of more robust and accurate person recognition systems.

\bibliographystyle{unsrt}  
\bibliography{references}  

\end{document}